\documentclass{article}

\usepackage{amsmath}
\usepackage{microtype}
\usepackage{graphicx}
\usepackage{subfigure}
\usepackage{booktabs} 
\usepackage{float}
\usepackage{hyperref}

\usepackage[accepted]{icml2019}

\icmltitlerunning{CS234 Final Report}

\begin{document}

\twocolumn[
\icmltitle{Compressed imitation learning}



\icmlsetsymbol{equal}{*}

\begin{icmlauthorlist}
\icmlauthor{Beicheng Lou}{equal,stan}
\icmlauthor{Nathan Zhao}{equal,stan}
\end{icmlauthorlist}

\icmlaffiliation{stan}{Stanford University}

\icmlcorrespondingauthor{Nathan Zhao}{nzz2102@stanford.edu}
\icmlcorrespondingauthor{Beicheng Lou}{lbc45123@stanford.edu}

\icmlkeywords{Machine Learning, ICML}

\vskip 0.3in
]



\printAffiliationsAndNotice{\icmlEqualContribution} 

\begin{abstract}
In analogy to compressed sensing, which allows sample-efficient signal reconstruction given prior knowledge of its sparsity in frequency domain, we propose to utilize policy simplicity (Occam's Razor) as a prior to enable sample-efficient imitation learning. We first demonstrated the feasibility of this scheme on linear case where state-value function can be sampled directly. We also extended the scheme to scenarios where only actions are visible and scenarios where the policy is obtained from nonlinear network. The method is benchmarked against behavior cloning and results in significantly higher scores with limited expert demonstrations. 
\end{abstract}

\section{Introduction}
\label{submission}
Reinforcement learning in high-dimensional problems can be very time-consuming when using traditional algorithms like Q-learning or SARSA. Learning from "expert" is thus desirable. This is the field of inverse reinforcement learning (IRL), which has seen many recent developments. In particular, a lot of effort has been levied at making IRL sample-efficient. In many realistic scenarios, an "expert" is not always available and/or "demonstrations" might be costly, which motivates techniques for efficient imitation through very limited observations. Some real world examples where this would apply include trading, where "expert" companies invest a lot in making their trades invisible and we could only observe a limited set of their trades. Other examples include training robots to perform various activities and training autonomous vehicles, where we hope to minimize the need for a human "expert" to demonstrate.

Furthermore, reconstructing the reward function is often an ill-posed problem, with many degenerate solutions. While many solutions have been proposed, such as maximum entropy reinforcement learning. However, once one determines the reward function, one still has to iterate through the training process to derive a policy from it, which is inefficient if the problem is high-dimensional. 

In order to address these shortcomings, we propose a novel algorithm which can accelerate an agent's Q-learning by using compressed sensing. It allows us to directly reconstruct the value function or action value function which trivially renders the policy.

\subsection{Compressed Sensing Formalism}

In a typical signal reconstruction problem, there is a well-known fact (Nyquist theorem) that one must sample a signal at the frequency of at least twice the highest frequency in the sample in order to get accurate reconstruction without aliasing. Compressed sensing is a regime of signal reconstruction where the number of samples is far below the Nyquist frequency \cite{Candes2008, CS}. In order to obtain high fidelity signal reconstruction in this regime, a few constraints can be introduced to exploit the prior knowledge. The most important of these is the notion that the signal has a "sparse" representation in some domain. For example, if we consider a single sinusoid ($sin(\omega t)$) in time, we know that it is a Dirac-delta function in the frequency domain. As a result, with just a few random samples in time domain of this signal even at the presence of noise one can, in principle, reconstruct it with arbitrarily small error.

The technical process of compressed sensing typically involves a minimization problem which exploits the sparsity of the signal in some domain under linear transformation (which can be generalized to nonlinear). Specifically, given a set of observations (in the dense domain) $y$ and a transformation operation $A$ which takes us from the sparse domain to the discrete domain, we want to reconstruct the signal $x$ to minimize its $L_0$ norm (i.e. maintain sparsity) subject to the constraint that it accurately reproduces all the observed samples:
\begin{equation}
    \min_{x} ||x||_0, \textrm{  s.t.  }  y = Ax
    \label{eq:CS}
\end{equation}

This is applicable when the signal has a low-dimensional structure and perfect reconstruction is feasible with very few samples. The measurement (sampling) matrix has to be incoherent and satisfy the restricted isometry property. In practice, we may relax the constraint as $|y-Ax|^2<\varepsilon$ or move it into the minimization objective as a term $|y-Ax|^2$. To make the optimization tractable, people often use  $L_1$ in the place of $L_0$, which is effectively the method of LASSO. Apart from gradient-based optimization, this problem can also be solved with iterative methods e.g. iterative hard/soft shrinkage.

Many imitation learning problems are in the regime where the model’s degrees of freedom outnumbers the expert’s demonstrations by orders, while there exists a concise representation of the expert’s policy (which makes imitation possible). The setting naturally motivates application of compressed sensing.

In direct analogy with the signal reconstruction task, we consider problem where we can get limited information about an expert and wish to  \cite{Candes2008, CS} reconstruct some aspect of the expert exploiting "sparsity" in some domain, which we will discuss later.

\subsection{Imitation Learning: Related Work}

Imitation learning aims to transfer an "expert's" knowledge of a particular task to another agent. This knowledge can come in the form of the expert's policy (behavior cloning, assuming that the agent and the expert follow the same set of dynamics) or more often (for robustness), in terms of an inferred reward function based on the observation of the expert's set of trajectories (inverse reinforcement learning).

As one of the simplest and most direct method of knowledge transfer, behavior cloning \cite{Bratko1995} involves the supervised learning task of fitting a function which most accurately maps a set of observed expert states to their respective action. This function then effectively serves as the agent's policy map. However, this can suffer from an inability to generalize if the model is given inputs outside the distribution that it is trained on (i.e. overfitting).

DAGGER, or dataset aggregation, overcomes the potential overfitting or lack of generalization of behavior cloning by simply providing an "online" platform whereby the agent can continuously query the expert for more labels to update the model \cite{Ross2011}. However, this clearly violates one of our starting premises of being data sparse. In fact, both BC and DAGGER can potentially require a significant number of training examples in order to work, which we hope to circumvent using compressed sensing.

Other related words include apprenticeship learning \cite{Abbeel2004}, which attempts to infer the reward function from the expert's demonstration. This method requires two steps, determining an approximation to the reward function, and then using this reward function to ultimately evaluate the action-value function. In our work, we will not require an inference of the reward function. In fact, we will not even require any knowledge of the reward function that the expert is operating with, only the expert's demonstrations of state and actions.

\section{Approach: Accelerating Imitation Learning with Compressed Sensing}
We now show that compressed sensing can provide a way to maximally transfer limited observations of an expert's state-action trajectories during Q-learning, both for a linear function approximator AND a nonlinear function approximator (neural net). 

\subsection{Level 1: Expert exposes Q(s,a)}
First, we must understand how we can formulate compressed sensing for inverse reinforcement learning. For simplicity, consider the action-value function $Q(s,a)$ with a linear function approximation:
\begin{equation}
    Q(s,a) = w^T \phi(s)
\end{equation}
where $\phi(s)$ is the feature vector for the state. In analogy with Eq. \ref{eq:CS}, we let $y \rightarrow Q(s,a)$, $w\rightarrow w$ and $x \rightarrow A $. Thus, if our weight matrix is sufficiently "sparse", Eq. \ref{eq:CS} actually directly applies to the reconstruction of $V$ with a limited set of actions. Similarly, this is applicable for a linear function approximation for $Q(s,a)$ (which is more practical as we can recover the policy from it).

The convex optimization problem is:

\vspace{-2ex}
\begin{equation}
\begin{aligned}
& \underset{w}{\text{minimize}}
& & ||w||_0 \\
& \text{subject to}
& & w_a^T \phi(s) = Q_{demo}(s,a) 
\end{aligned}
\label{eq:sa}
\end{equation}
\vspace{-2ex}

We note that solving this problem can be done online or offline in the batch setting. In the former case, if a subsample of the expert's actions do not yield an adequate reconstruction, the procedure can be re-run again to obtain a better result. In the latter case, one must hope that the batch of data provided is diverse and contains sufficient information to provide a solid reconstruction.

\subsection{Level 2: Sampling only Expert's states and actions}
Of course, one downside of the formulation presented in the previous section is that in practice, an expert's value function is not directly observable. We only have observation of the expert's states, actions and rewards (to reconstruct either $V$ or $S$, we would need further knowledge of the reward function at a minimum. However, the formalism fits very nicely into the framework of CS and only requires that we can demonstrate that $w$ is sparse and demonstrations satisfy the restricted isometry property \cite{Candes2008}.

However, we can also construct a CS problem using just the observed information from the expert's states and actions. For every state-action pair (s,a) observed in the demonstration:
\vspace{-1ex}
\begin{equation}
\begin{aligned}
& \underset{w}{\text{minimize}}
& & ||w||_0 \\
& \text{subject to}
& & w_a^T \phi(s) > w_{a'}^T \phi(s) \quad \forall{a'}
\end{aligned}
\label{eq:sa}
\end{equation}
\vspace{-1ex}

More specifically, in the cartpole example, our action space has cardinality 2, so we construct a feature vector for each state $s_i$ demonstrated by the expert where action $a_i$ is taken:
\begin{equation}
    A_i = \begin{cases}
        [\phi(s_i)^T \quad -\phi(s_i)^T] & a_i=0 \\
        [-\phi(s_i)^T \quad \phi(s_i)^T]& a_i=1
    \end{cases}
\end{equation}

With the weight vectors concatenated into $x^T = [w_0^T \quad w_1^T]$,
equation \ref{eq:sa} can be brought to a form similar to \ref{eq:CS} with constraint $Ax \ge 0$.

In practice, and as will be discussed later, exact optimization on the $L_0$ norm is an NP-hard problem and quickly becomes intractable even for modest problem sizes. As a result, we consider an L1-norm problem: 
\vspace{-1ex}
\begin{equation}
\begin{aligned}
& \underset{w}{\text{minimize}}
& & ||w||_1 + \lambda ||w - w_{\text{target}}||_2\\
& \text{subject to}
& & w_a^T \phi(s) > w_{a'}^T \phi(s) - \epsilon \quad \forall{a'}
\end{aligned}
\label{eq:sa_reg}
\end{equation}
\vspace{-1ex}

The regularization term is now added because of the nature of the constraint in Eq. \ref{eq:sa_reg}. $w = 0$ trivially satisfies this constraint so that optimizing the L1 norm will also trivially just force the weights to be zero. For the linear case, the choice of a random target weight is sufficient to prevent this trivial solution from being selected. 

The $\epsilon$ is another hyperparameter which relaxes the condition that $Q(s,a') \geq Q(s,a)$ for all other $a$. This hyperparameter is useful in cases where the expert is noisy or is not an "optimal" expert. While the presence of two hyperparameters that we have to scan may pose a problem in a data-sparse problem, it isn't because they require only data that the agent collects and we may assume that the agent can easily collect data from the environment.

One interesting observation to note is that because only a finite number of samples are provided by the expert, the $Q(s,a)$ that is reconstructed may not be unique. Intrinsically, there are just multiple possible $Q(s,a)$ which could output the correct set of states and actions from the expert.

\subsection{Level 3: Application to Deep Q Networks}

Function approximation without the use of deep neural network is difficult to tune in practice, thus it is desirable to extend the method to imitation learning on Deeqp Q Network (DQN) or deep policy network. One challenge there is the use of nonlinear measurements and nonlinear constraints, which conflicts the CS methodology in nature. Remedies typically relax the contraint into regularization terms, which then transforms the challenge into the ambiguity of the line between compressed imitation learning and imitation learning with a clever regularization (that captures the same prior).

%
\begin{figure}[h]
    \centering
    \includegraphics[width = 2 in]{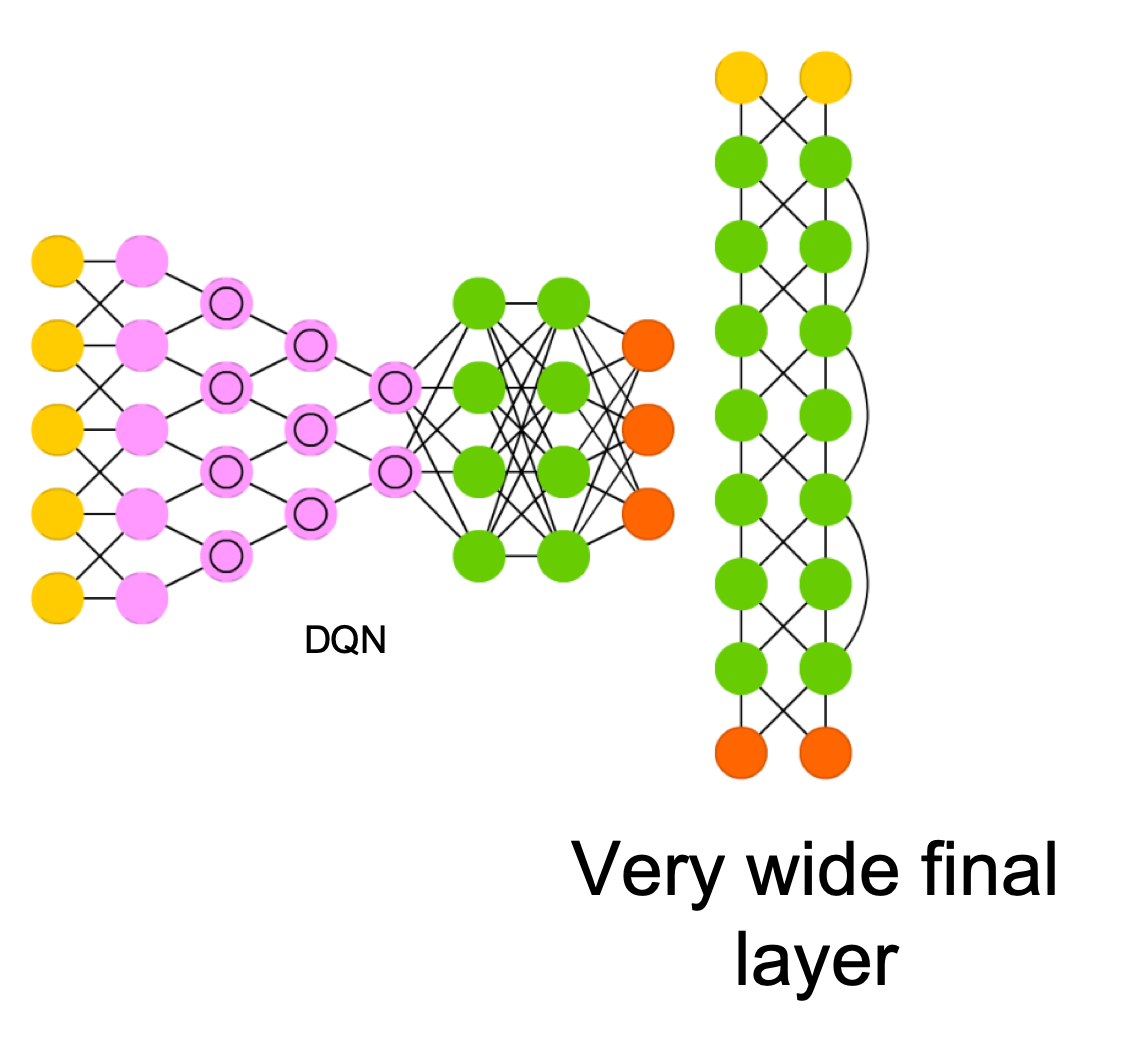}
    \caption{Schematic of the DQN architecture for compressed sensing. It is essentially a typical DQN but the last layer is artificially a wide layer and trained with dropout.}
    \label{fig:my_label}
\end{figure}

A Deep Q network (DQN) can be viewed as a combination of preparing a highly nonlinear feature vector (up to second last layer in DQN) and applying the correct weights on the feature for final output (the final layer in DQN), which allows us to apply compressed sensing on the last layer of the network in a linear fashion. Our network architecture employs an extremely wide last layer which we train with dropout so that the CS applies.

We denote the DQN as a function:
\begin{equation}
    Q(\phi(s); w_1, ...w_n).
\end{equation}
The compressed sensing problem is:
\vspace{-1ex}
\begin{equation}
\begin{aligned}
& \underset{w_n}{\text{minimize}}
 ||w_n||_{nuc} \\
& \text{subject to } \\
& Q(\phi(s), a; w_1, \cdots, w_n) >\\
& Q(\phi(s), a'; w_1, ..., w_n) \quad \forall{a'}
\end{aligned}
\label{eq:sa}
\end{equation}
\vspace{-1ex}

To interpret this properly, 
\begin{equation}
    Q(\phi(s); w_1, ...w_n) = w_n^T (f(\phi(s)).
\end{equation}

Essentially, in the DQN we treat the initial layers of the network as a special 'featurizer' of $\phi(s)$. The final layer is simply a linear layer which maps $f(\phi(s))$ to $Q(s,a)$. In more complex problems, this procedure can be interleaved with the normal training process of DQN, where the normal training improves the whole network (especially the featurizer) gradually and CS boosts the final layer weights once in a while.


\section{Theory}
In general, solving Eq. \ref{eq:CS} is an NP-hard problem due to the 0-norm. There are a few relaxations which can still guarantee good solutions, as discussed in Section~1.1.




One of the key requirements of compressed sensing is the Restricted Isometry Property (RIP), which is a quantitative realization of the conditions that the measurement and representation bases are incoherent.

\begin{equation}
    (1-\delta_k)||x||_2^2\leq||Ax||_2^2 \leq (1+\delta_k)||x||_2^2 
    \label{eq:rip}
\end{equation}

Eq. \ref{eq:rip} requires that the projection of $A$ (the reconstruction matrix) preserves distance between signals.
An equivalent statement of the RIP is that the eigenvalues of $A^TA$ lie in the interval $[1-\delta_k, 1+\delta_k]$. It has been proven that RIP is naturally satisfied for random matrices \cite{RIP-random}.

In the case of Q-learning, the sensing matrix $A$, consists of $\phi(s)$, which means that the burden of determining whether CS will work lies in analyzing $\phi(s)$ for a given RL problem.
For us, $A$ is a stack of high-dimensional feature vectors $\phi(s)$ for all states s sampled from expert, which is approximately a random matrix, despite some correlation in the feature vectors that slightly increases difficulty but gradually diminishes as dimension increases.

In general, there are several important considerations to understand that make the application of RIP and similar properties used in CS a bit more nuanced when using it for RL. The biggest one is the fact that the transformation from sparse basis to measurement basis $y=Ax$ is not "unique". $A$ can be different (or even inherently stochastic) depending on what demonstrations the expert provides. Moreover, it also means that there is not a unique $x$ which would solve the problem. 

Additionally, for Q-learning, the fidelity in which $Ax$ matches $y$ is also not the same as that of a typical CS problem. Since in the reinforcement learning problem, the performance of the agent is predicated on selecting the correct action rather than predicting the right $Q(s,a)$ value, we care more that the resultant policy $\pi(a) = \text{argmax}_{a'}Q(s,a')$ matches the expert. Namely, the exact value of the reconstruction does not matter as much as the sign of it.



%

\section{Results}
\subsection{Problem and Data Setup}
Our first benchmarks are using the OpenAI gym's Cartpole. In Cartpole, the state of the system is given by a four dimensional vector which contains the cart position, cart velocity, pole angular position, and pole angular velocity respectively. As the state space of this is relatively small, we artificially expand it using a union of five different nonlinear gaussian kernels. The action space for the cartpole is discrete consisting of two actions corresponding to moving the cart left or right. We denote these two actions with $a = 0 $ and $a = 1$.

We use sklearn's radial basis function (rbf) sampler, which generates a feature map of an RBF kernel by Monte Carlo approximation of its Fourier transform. To fit the sampler, we generate a set of 20000 random observation samples from the environment's state space to fit the featurizer. With five different kernels, each 100 components each, the resultant feature dimension for this problem is 500 instead of just 4. In doing so, we also guarantee that the weight vector for this problem will have some degree of "sparsity".

In order to train an "expert" on this problem, we use typical Q-learning. The baseline Q-learning to train our agent takes around 10000 or more episodes to yield perfect performance. With each episode containing up to 200 different state-action pairs, this amounts to almost 1 million d\vspace{-1ex}
\begin{equation}
\begin{aligned}
& \underset{w}{\text{minimize}}
& & ||w||_{nuc} \\
& \text{subject to}
& & w_a^T \phi(s) > w_{a'}^T \phi(s) \quad \forall{a'}
\end{aligned}
\label{eq:sa}
\end{equation}
\vspace{-1ex}ifferent state-action episodes needed to converge (recall that we are expanding the state dimension so this increases the amount of data needed).

\begin{figure}[h]
    \centering
    \includegraphics[width = 3 in]{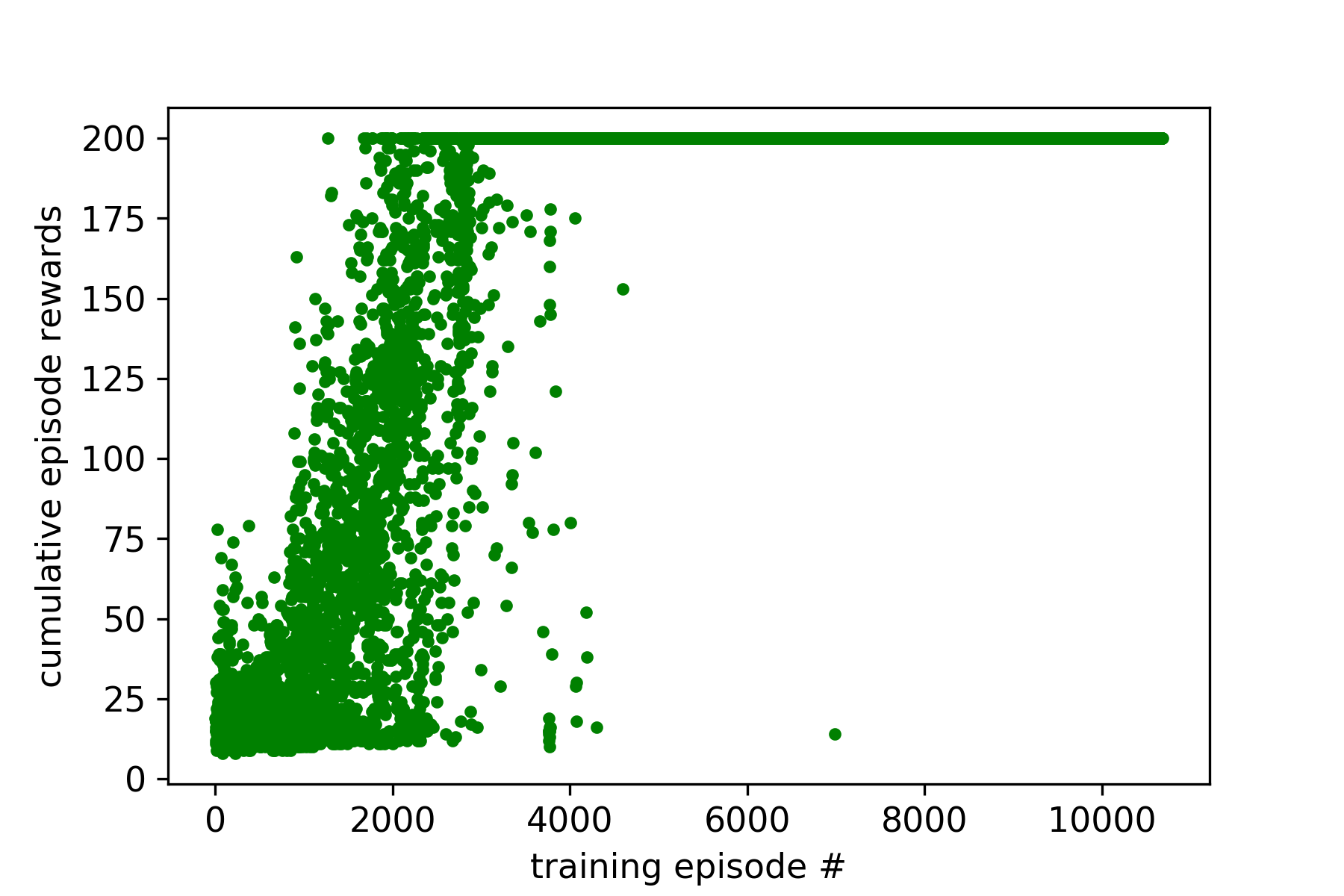}
    \caption{Training curve for Q-learning for our "expert" on our 500-dimensional cartpole verifying that the expert can still be trained to "perfect" performance on this high dimensional space.}
    \label{fig:q-learning}
\end{figure}



\subsection{Level 1: Expert exposes Q(s,a)}

With our expert generated, we then provide a limited set of examples from which we learn from.  In the level 1 case, we again assume that the expert actually provides its $Q(s,a)$ function values, which is heavily contrived, but serves to illustrate that the compressed sensing works. In the following results, we use ONLY \textbf{21 examples}, which represents a number of examples that's at least 2 orders of magnitude smaller than what we trained on. First, using the expert's $Q(s,a)$ we show the reconstruction performance.
\begin{figure}[h]
    \centering
    \includegraphics[width=3.2 in]{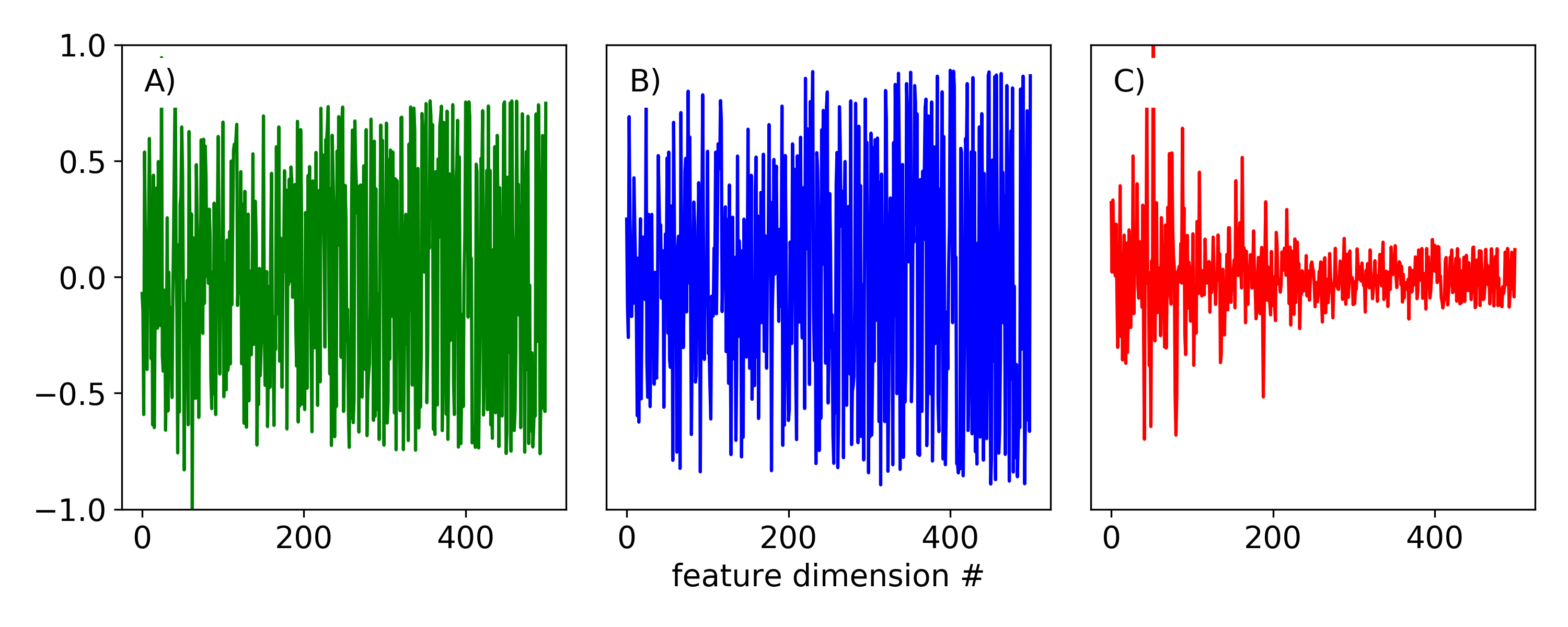}
    \caption{a) Reconstruction of the weights of the action-value function using compressed sensing b) original weights obtained from an expert and c) the difference of b) and c) which gives an approximate estimate of the error between the reconstruction and the true expert's weight.}
    \label{fig:reconstructed_weights}
\end{figure}

While a side by side comparison of the original weights and the reconstructed weights does not appear to show a match, looking at the difference is much more instructive. The average absolute value of the error in the reconstructed weights is just 10\%. Even given the relatively sizeable errors however in the weights, we can see that the reconstructed agent's performance is actually still quite good.

\begin{figure}[H]
    \centering
    \includegraphics[width = 3. in]{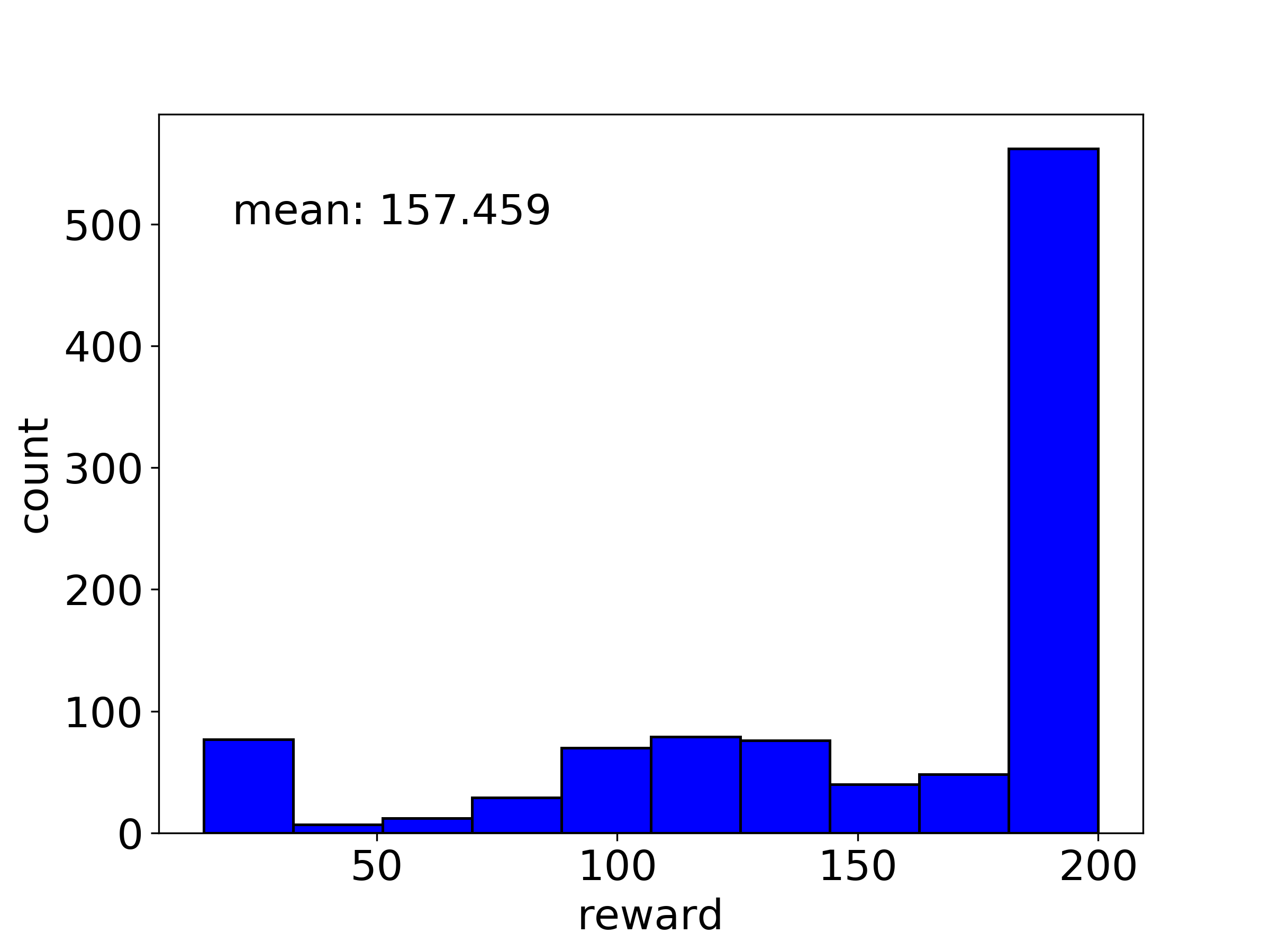}
    \caption{Distribution of test rewards using the reconstructed weight (shown in Fig. 3) from 21 expert demos. The average of the test rewards is approximately 157.}
    \label{fig:my_label}
\end{figure}


Even given the error, the reason why the reconstruction is relatively good can be better elucidated once we realize again that it is whether or not the reconstructed action matches between the reconstructed agent and our expert. In Fig. \ref{fig:Q_difference}, we plot the difference of $Q(s,1)-Q(s,0)$ for the expert versus the agent. This plot tells us that if the differences are more correlated then our reconstructed agent tends to pick the same action as the original expert. And indeed we do see a strong correlation.

\begin{figure}[h]
    \centering
    \includegraphics[width = 2 in]{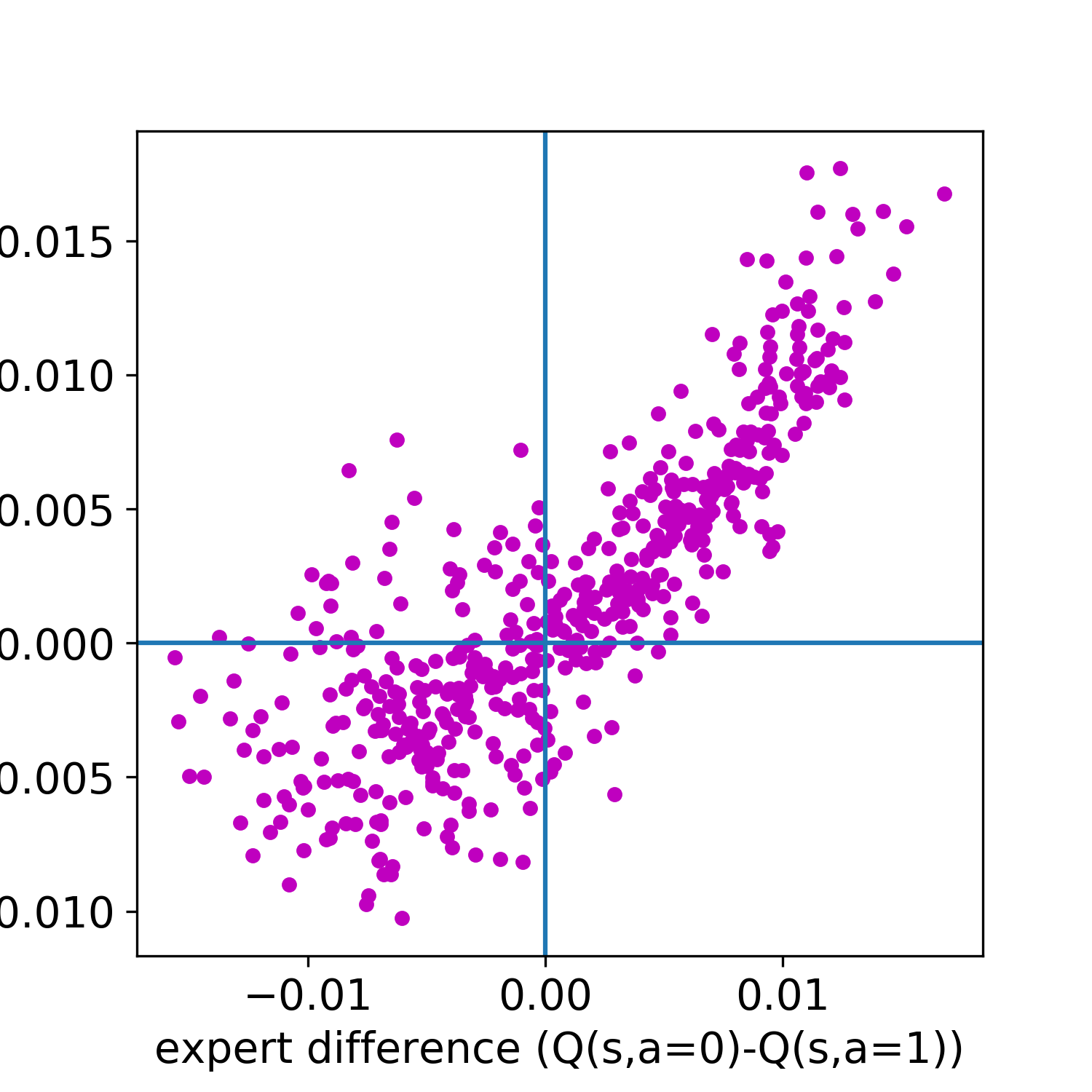}
    \caption{Plot of the difference of Q(s,0) - Q(s,1) of the expert (x-axis) versus that of the reconstructed level 1 agent (y-axis). The correlation coefficient is ~0.8}
    \label{fig:Q_difference}
\end{figure}


\subsection{Level 2: Expert exposes only state-action pairs}
Now we consider the problem of reconstruction with only the state-action pairs visible $(s,a)$ provided by the expert as opposed to the Q(s,a) values. This kind of expert exposure is reasonable.

First, we need to identify what to choose as an appropriate $w_{target}$ in Eq. \ref{eq:sa_reg}. Empirically, we find that picking a random $w_{target}$ is sufficient to generate good reconstructions.

\begin{figure}[H]
    \centering
    \includegraphics[width = 3.2 in]{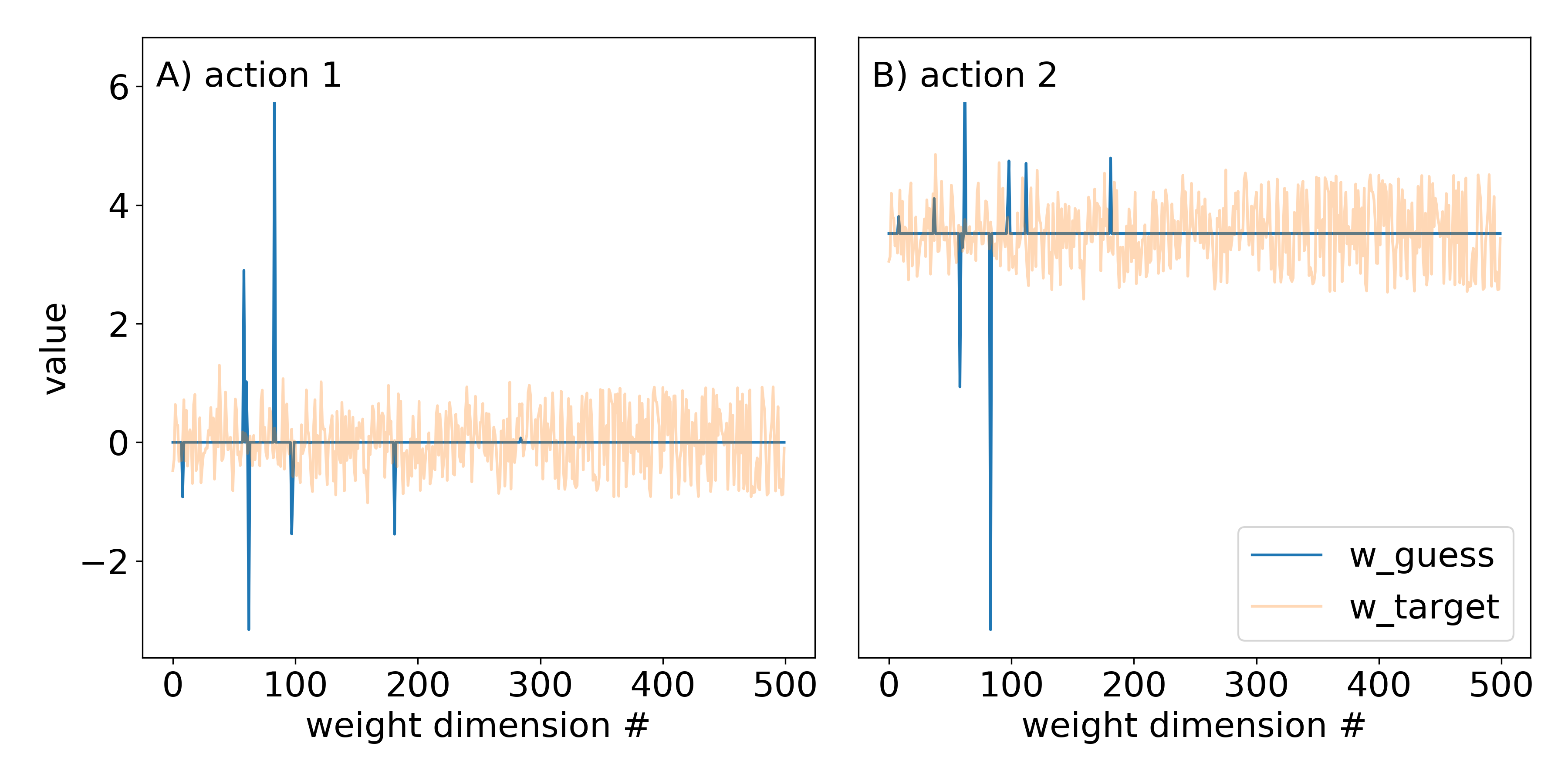}
    \caption{A) reconstructed weight for action 1 (in blue) and the corresponding random $w_{target}$ used. B) The analogous set of  results for action 2. }
    \label{fig:level2_reconstruction}
\end{figure}

In Fig. \ref{fig:level2_reconstruction}, we see that the reconstructed weights are actually k-sparse, in contrast to level 1. the orange lines indicate the random $w_{target}$ used. Being able to achieve the k-sparsity here is critical as the agent is able to effectively identify the few features in the 500 dimensional space that are actually useful. What is truly remarkable however is that using this formulation, we can actually get the test performance of the reconstructed agent is nearly perfect!  

Below, we show the analogous performance metric as shown in Fig. \ref{fig:q-learning}. As expected, for perfect performance, the agreement between the expert Q difference and the agent's Q difference is much tighter than in level 1.

\begin{figure}[h]
    \centering
    \includegraphics[width = 3.2 in]{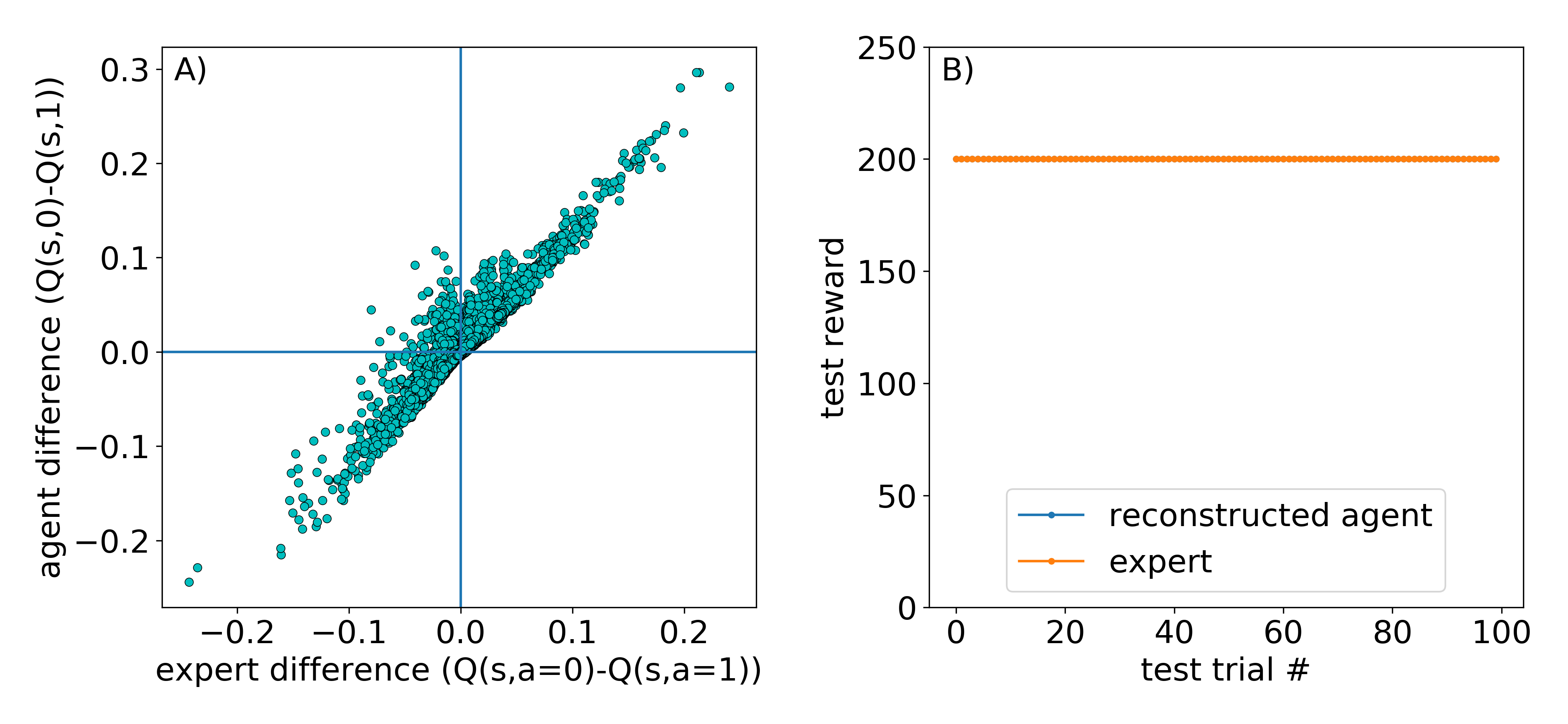}
    \caption{Plot of the difference of Q(s,0) - Q(s,1) of the expert (x-axis) versus that of the reconstructed level 2 agent (y-axis). The correlation coefficient is 0.73. }
    \label{fig:Q_difference}
\end{figure}

In effect, by receiving only 21 demonstrations of the expert, compressed sensing is already sufficient to give the agent an almost perfect image of the expert's policy for the cartpole.

\begin{figure}[H]
    \centering
    \includegraphics[width = 2 in]{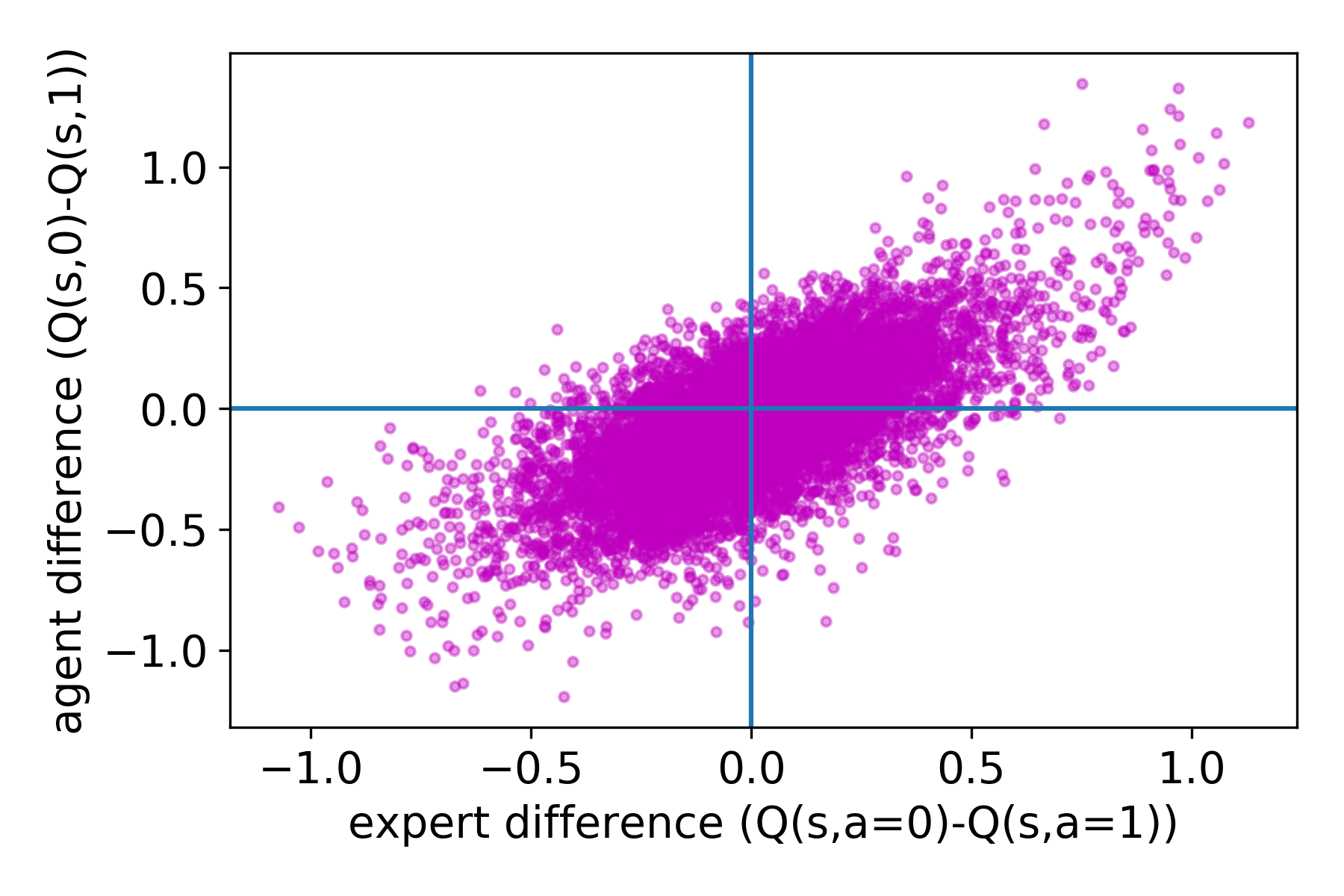}
    \caption{Plot of the difference of Q(s,0) - Q(s,1) of the expert (x-axis) versus that of the reconstructed agent (y-axis) for states sampled uniformly across the entire spectrum of the state space. Again we see that our reconstructed agent matches the expert relatively closely for the majority of states. }
    \label{fig:Q_difference_whole_state}
\end{figure}

Moreover, we can evaluate how well the agent performs compared to the expert across the entire state space (as oppose to the restricted set of states where the cartpole is mostly balanced). This is shown in Fig. \ref{fig:Q_difference_whole_state}.

\subsection{Level 3: DQN}
\begin{figure}[H]
    \centering
    \includegraphics[width=3.5 in]{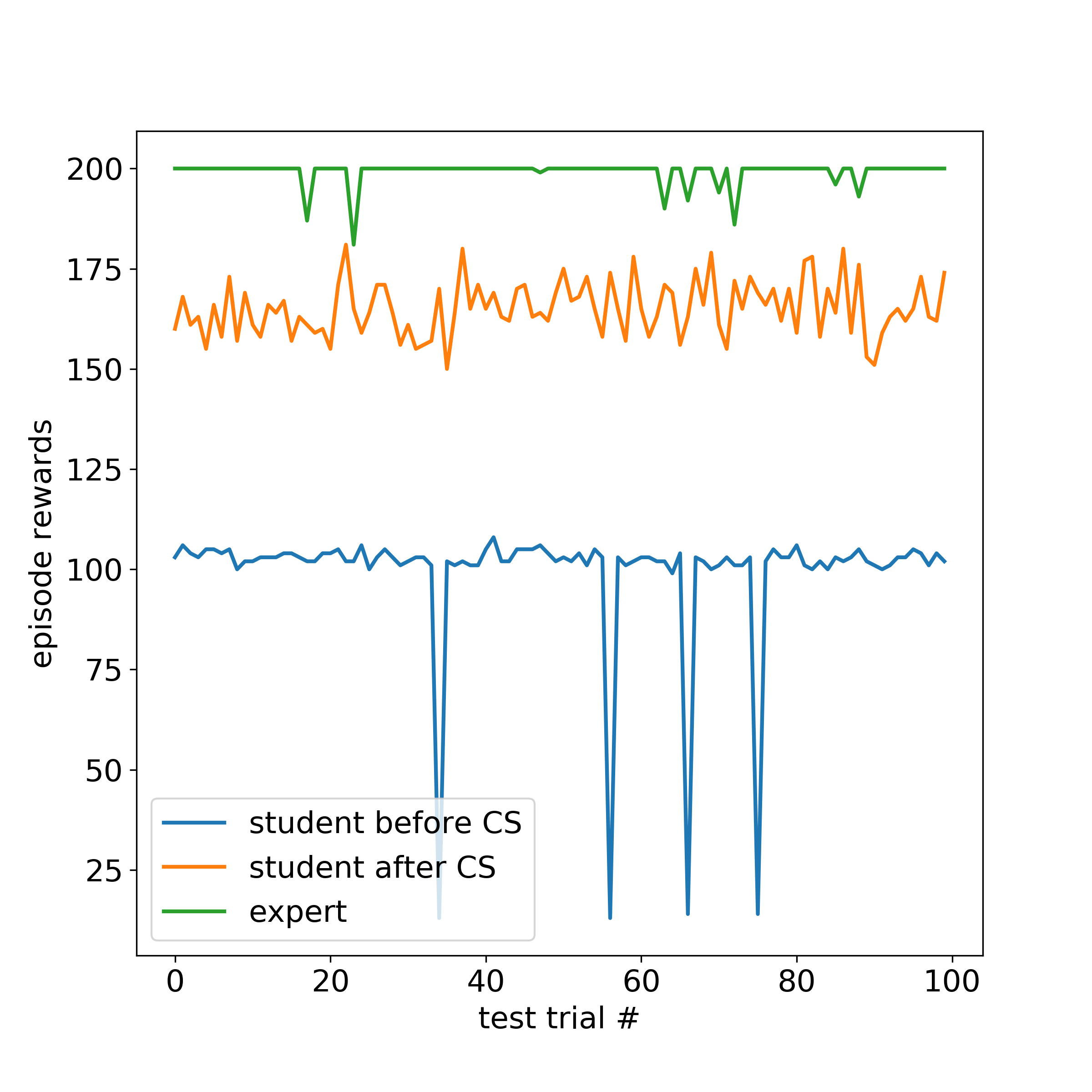}
    \caption{Performance of DQN with CS on the last layer. The blue line represents our partially trained DQN "student", the green line represents our fully trained expert, and the orange line represents the performance of our student after being given 20 demonstrations from the expert.}
    \label{fig:dqn_performance}
\end{figure}

Finally, we demonstrate the success of our formalism using a nonlinear function approximator.
We start with a "student" DQN. The architecture of the DQN is like any typical neural net but we expand the last layer to be a very wide and train it with dropout. WE "pretrain" this net for a relatively small number of iterations, which is insufficient to get good performance, as shown by the blue line in Fig. \ref{fig:dqn_performance}. At that point, we expose some expert state action pairs. These state action pairs are used to "sparsiy" the weights of the last layer. Then we re-evaluate the performance of our "student" network to see if there is improvement. The purpose of the pretraining is to acquire a $w_{target}$.

\begin{figure}[H]
    \centering
    \includegraphics[width = 3 in]{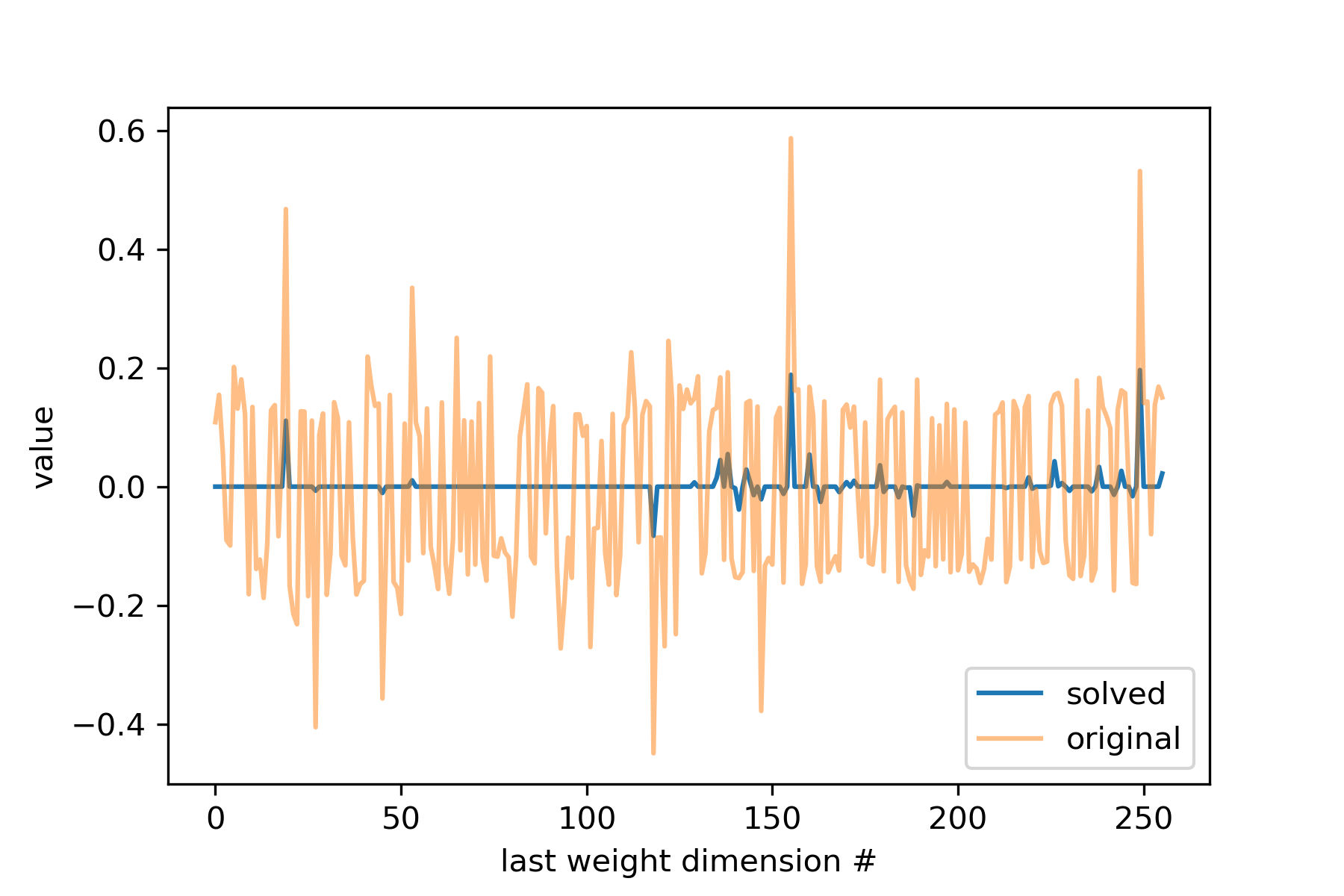}
    \caption{Reconstruction of the last layer weights using compressed sensing (blue) versus the last layer weights of the trained expert (orange).}
    \label{fig:dqn_last_layer}
\end{figure}

In Fig. \ref{fig:dqn_last_layer}, we can see again, just as in the linear case, the reconstruction process substantially sparsifies the weights in the last layer.

In Fig. \ref{fig:dqn_performance}, the blue line represents the average test performance of the student network after 1000 iterations. The orange line shows the test performance after doing CS on the last layer with 20 expert examples. And we show our baseline expert's performance (a fully trained dqn for 20000 epochs), which is near-perfect.

\section{Benchmarks: Behavior Cloning}
While our results above are already significant as they provide a way to vastly accelerate Q-learning by directly integrating the CS step into the Q-learning algorithm, we compare our technique to behavior cloning to directly compare how much of an advantage we gain in number of training examples.

\begin{figure}[H]
    \centering
    \includegraphics[width = 2.75 in]{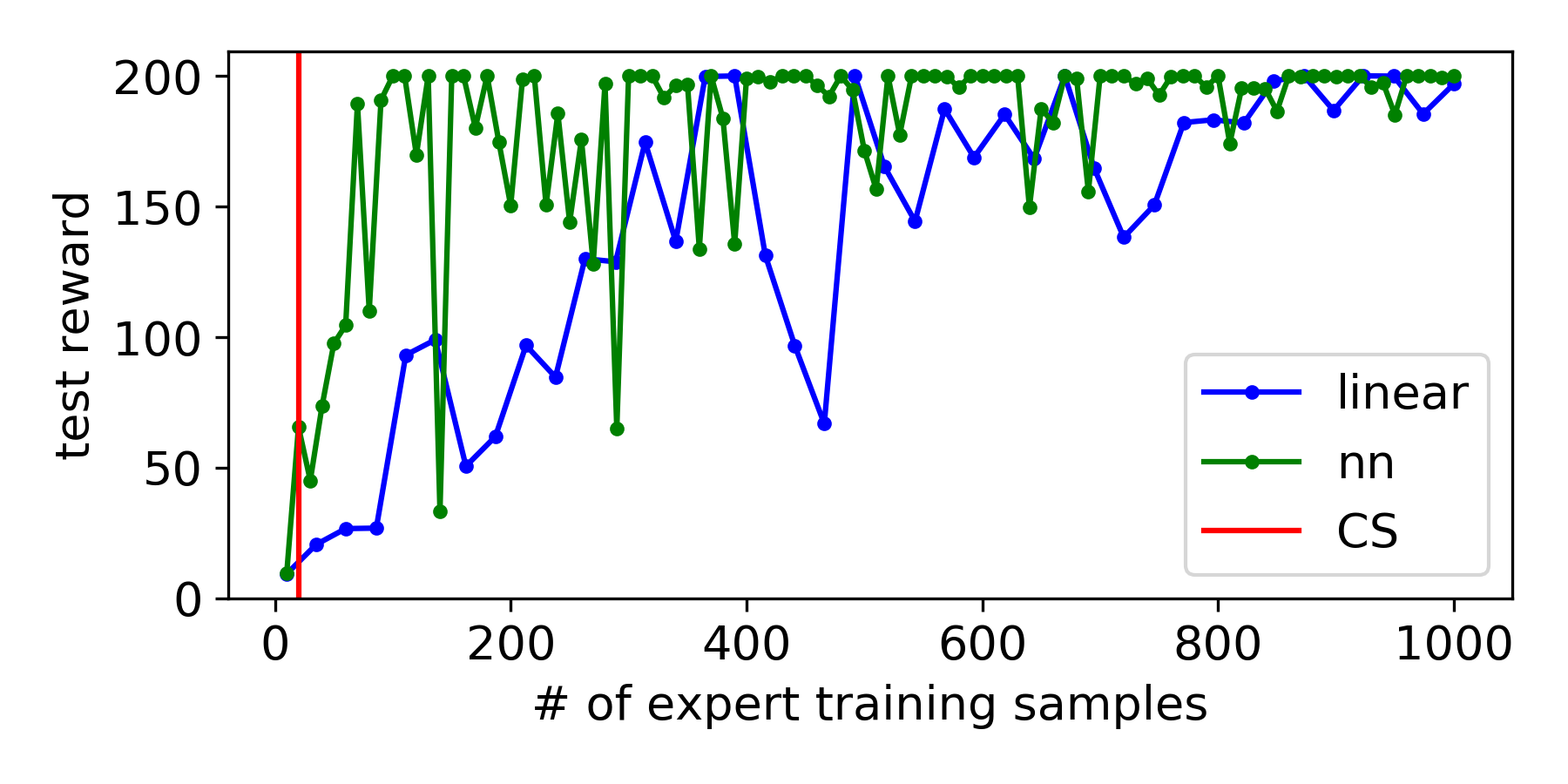}
    \caption{The reward (in test) of a behavior clone trained on 10-2000 samples. Roughly more than 1000 samples are needed to get consistent perfect performance (reach time step 200 without debalancing the pole).}
    \label{fig:behavior_cloning}
\end{figure}

Clearly, with behavior cloning, one needs well over 500 examples to get consistent PERFECT performance in test, which gives direct CS at least an order of magnitude improvement over behavior cloning.

\section{Conclusion}
We have demonstrated a novel procedure which allows us to apply compressed sensing directly to existing reinforcement learning algorithms. For the appropriate problem domains where the feature space is high dimensional, our technique can substantially accelerate the imitation process. Moreover, it requires strictly only that the expert demonstrates what their action is for a given state. This is empirically demonstrated on three different levels with increasing generality: the linear QFA case where expert exposes Q, the linear QFA case where expert only shows state-action pairs and the DQN case where expert only shows state-action pairs. Our method gives the agent a significant boost in performance with very few expert samples for all three cases.

Future work includes:
\begin{itemize}
    \item Generalize the restricted isometry property for the scenario where the reconstruction goal is relaxed (as we only care about the relative state-action values)
    \item Extend the scheme to DPN. This is challenging since the expert's probability of actions cannot be directly observed, thus aiming to best reconstruct the expert's demonstrations will have a variational flavor where we try to maximize the likelihood of the demonstrated actions, which is incompatible with the one-shot nature of compressed sensing.
    \item Apply the method on more complex problems. The method's advantages will be amplified in high-dimensional problems, but the training time will also be longer and exceeds the scope of this project.
    \item Generalize the prior to be encoded by a generative network. Here we require the weight vector to be sparse in its own domain, whereas in principle, it only needs to have a concise/compact representation. That suggests possibility of encoding the weight vector by a generative model and put the sparsity constraint on the input to the generative model.
\end{itemize}






\nocite{langley00}

\bibliography{example_paper}
\bibliographystyle{icml2019}

\end{document}